\documentclass{article}
\usepackage{ijcai}

\usepackage{times}
\usepackage{soul}
\usepackage{url}
\usepackage[hidelinks]{hyperref}
\usepackage[utf8]{inputenc}
\usepackage[small]{caption}
\usepackage{graphicx}
\usepackage{amsmath}
\usepackage{amsthm}
\usepackage{booktabs}
\usepackage{algorithm}
\usepackage{algorithmic}
\usepackage{color}
\usepackage{pstricks}
\usepackage{xcolor}
\usepackage{enumerate}
\usepackage{arydshln}
\usepackage{subfig}
\usepackage{pifont}
\usepackage{xspace}
\usepackage{latexsym}
\usepackage{float}
\usepackage{amsfonts}

\usepackage{multirow}

\usepackage{times}
\usepackage{latexsym}
\usepackage{float}

\usepackage[T1]{fontenc}

\usepackage[utf8]{inputenc}

\usepackage{microtype}
\usepackage{amsmath, amssymb}
\usepackage{booktabs}
\usepackage{graphicx}
\usepackage{multirow}
\usepackage{subfloat}
\usepackage{enumitem}

\title{\modelname: Sales Script Extraction and Analysis from Sales Chatlog}

\author{
Hua Liang$^{*1}$
\and
Tianyu Liu$^{*1}$\and
Peiyi Wang$^{^\dag2}$\and
Mengliang Rao$^1$ \And
Yunbo Cao$^1$
\affiliations
$^*$Equal Contribution 
$^\dag$Work during Internship at Tencent\\
$^1$Tencent Cloud Xiaowei
$^2$Institute of Computational Linguistics, Peking University\\
\emails
\{suvedoliang, rogertyliu, peiyiwang, sekarao, yunbocao\}@tencent.com
}

\begin{document}
\newcommand{\modelname}{\textsc{SmartSales}\xspace}

\maketitle

\begin{abstract}
In modern sales applications, automatic script extraction and management greatly decrease the need for human labor to collect the winning sales scripts, which largely boost the success rate for sales and can be shared across the sales teams.  
In this work, we present the \modelname system to serve both the sales representatives and managers to attain the sales insights from the large-scale sales chatlog. 
\modelname consists of three modules: 
1) \textit{Customer frequently asked questions (FAQ) extraction} aims to enrich the FAQ knowledge base by harvesting high quality customer question-answer pairs from the chatlog. 
2) \textit{Customer objection response} assists the salespeople to figure out the typical customer objections and corresponding winning sales scripts, as well as search for proper sales responses for a certain customer objection.
3) \textit{Sales manager dashboard} helps sales managers to monitor whether a specific sales representative or team follows the sales standard operating procedures (SOP).
The proposed prototype system is empowered by the state-of-the-art conversational intelligence techniques and has been running on the Tencent Cloud to serve the sales teams from several different areas. 

\end{abstract}

\section{Introduction}
In the sales teams, some sales representatives consistently attain or exceed the sales goals while others do not. 
The conventional routine of escalating the performance of sales teams is to figure out what makes good sales stand out by manually analyzing hundreds or thousands of human-to-human conversation chatlog between salespeople and customers, and then summarize the winning sales script or other sales tips for sales team training \cite{leong1989knowledge,leigh1989mapping,humphrey1994cognitive}.
However, this labor intensive procedure requires the participation of sales experts and tedious human analysis, which makes it undesirable for modern fast-paced, high-growth sales teams.

Recent technological advances in natural language processing (NLP) have made it possible to induce the skeleton or key semantics of the human conversations in an automatic fashion \cite{yu-etal-2020-dialogue,gliwa-etal-2019-samsum,sun-etal-2019-dream}. 
Compared with recognizing the winning sales scripts and analyzing the large-scale sales chatlog through human experts, the automatic winning sales scripts mining and analyses empowered by advanced NLP techniques would largely accelerate the performance improvement of sales teams.
To this end, we build an easy-to-use system named \modelname that is capable of handling hundreds of thousands or even millions of sales chatlog and assists the both sales representatives and managers to figure out the best practises to increase the sales success rate.
\modelname is a web application that consists of three modules: \textit{Customer FAQ extraction}, \textit{Customer objection response mining and search}, and \textit{Sales manager dashboard}.

\modelname has been deployed on the Tencent Cloud and can also be seamlessly embedded in the existing customer relationship management (CRM) systems.
We verify the performance of \modelname according to the feedback from an online education sales teams and an automobile sales teams.

\section{System Architecture}

\begin{figure*}[h]
    \centering
    \includegraphics[width=1.0\linewidth]{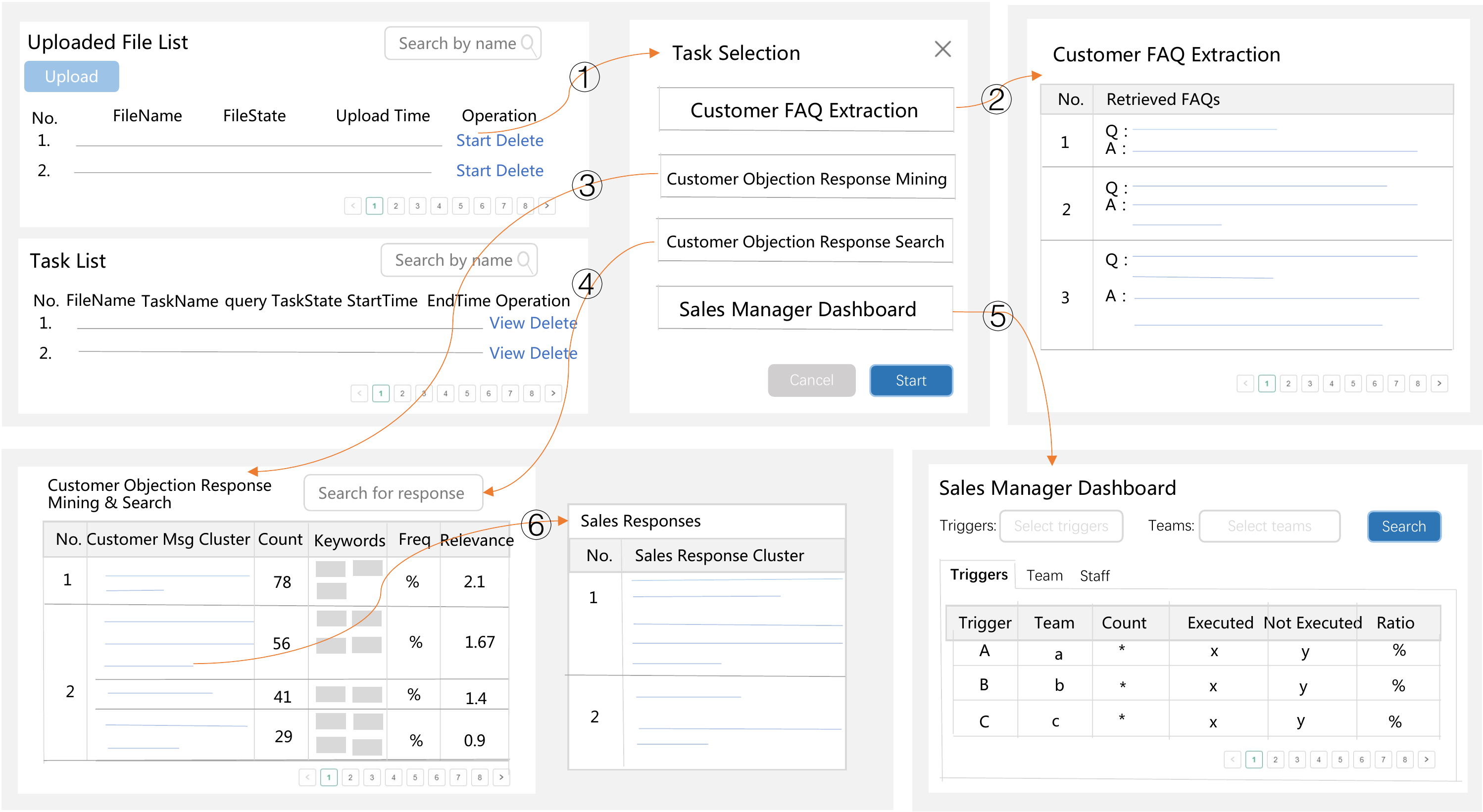}
    \caption{The illustration for the user interface of \modelname (the UI elements are translated to English from Chinese).  The users start from an entrance page ({\large \textcircled{\small 1}}) that includes file uploading and task selection ({\large \textcircled{\small 2}}-{\large \textcircled{\small 5}}). The selected tasks correspond to the different modules of \modelname, which are elaborated in Sec \ref{sec:interface}. }
    \label{fig:interface}
\end{figure*}

\subsection{Modules}
\paragraph{Customer FAQ Extraction}
As a replacement for manual FAQ accumulation, this module offers a much easier and more efficient way for sales teams. The question-answer pairs retrieval followed by human post-checking would be one of the cornerstones for knowledge acquisition as new customers emerge. According to the real customer feedback, this module reduce the human cost by up to 97\%, 87\%  during knowledge base cold-start for the sales teams of an international express company and an online education company respectively.

\begin{figure}[t]
	\centering
    \includegraphics[width=1.0\linewidth]{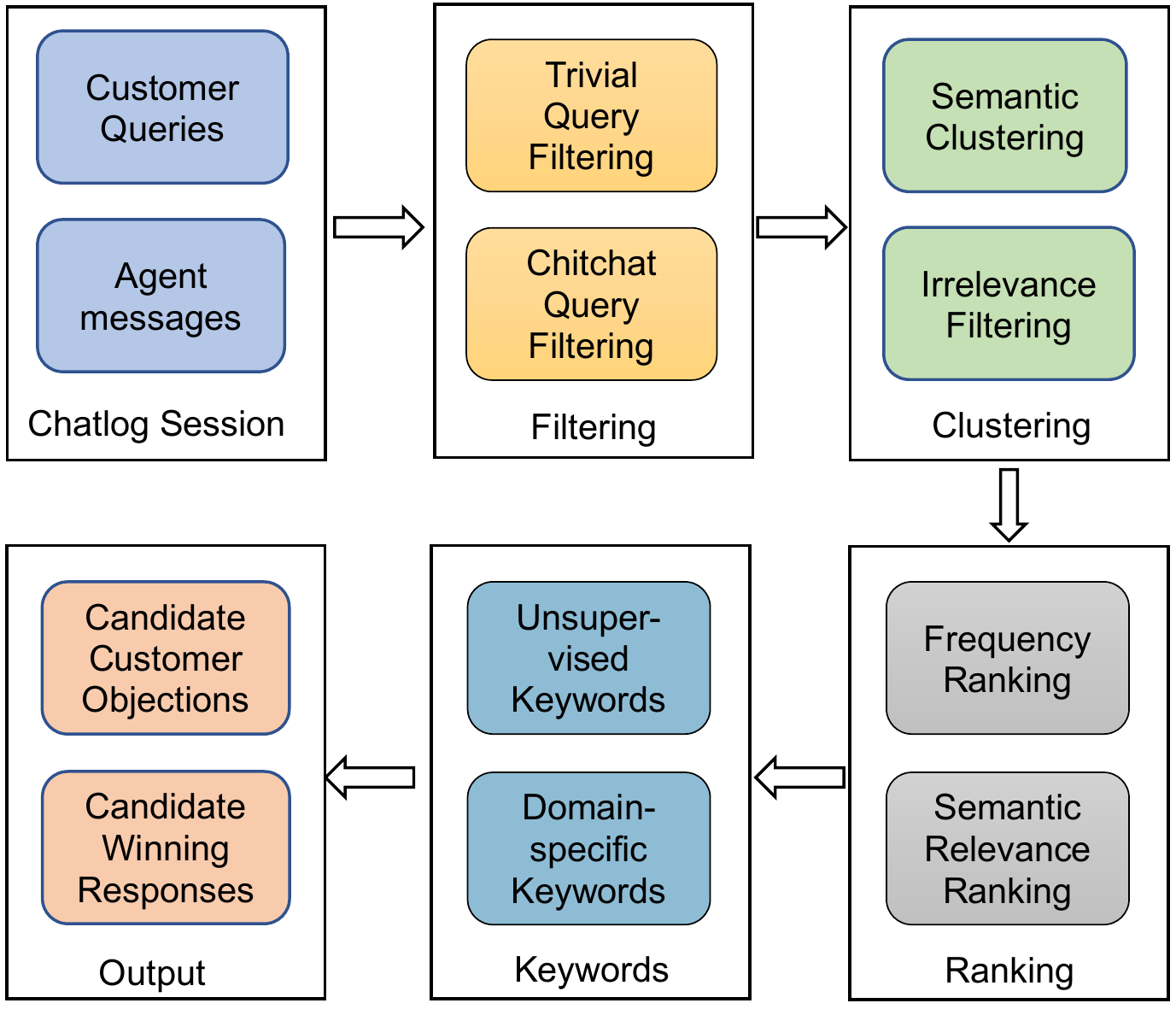}
	\caption{The NLP pipeline for \textit{customer objection response} module.}
	\label{fig:obj_arch}
\end{figure}

\paragraph{Customer Objection Response}
Customer objections are the concerns that a prospect has which cause them to hesitate (at best) and abandon (at worst) a purchase.
Even professionals might feel nervous while they find themselves facing the customer objections. 
One of the common resolutions is to highlight the typical customer objections that come up again and again by going through the chatlog and then create a plan to answer them.
However, the sales chatlog would grow rapidly for a large sales group which has vast prospective customers, and the customer objections might change in the different stages of the sales circle.
The proposed module would serve as an assistant for the salespeople to determine typical customer objections and gain actionable insights from good responses in the periodical sales retrospectives or performance reviews. 
As illustrated in Fig \ref{fig:obj_arch},
after filtering trivial customer messages, semantically similar customer queries and corresponding responses from sales are assembled in the same cluster. The clusters of customer queries are ordered by frequency and semantic relevance. For each cluster, we offer keywords that highlights the core semantics of customer queries.
The sales representatives could figure out the typical customer objections and the best practises to respond, or search for a successful sales response for a particular customer objection with the proposed module.


\paragraph{Sales Manager Dashboard}
In order to increase the sales performance, the sales manager should understand the pros and cons of their crew in order to develop the training programs for the sales staffs.
The proposed module helps the managers to overview the sales performance with a ``\textit{query trigger-response spotlight}'' paradigm.
The paradigm consists of a set of pre-defined rules that specify the sales standard operating procedures (SOP), i.e. ``what the sales should do in the given situation''. For example,
\begin{itemize}
    \item When a customer hesitates due to the affordability, the sales should mention payment policy such as ``pay by installments''.
    \item While facing a new customer, the sales should advertise for the best-seller productions.
\end{itemize}
The \textit{customer query triggers} and \textit{sales response spotlights} are formulated as a set of rules, keyword matching and query intention classification models, which are tailored for different sales groups.

We verify the module functionality and effectiveness of \textit{customer objection response} and \textit{{sales manager dashboard}} with the chatlog from an automobile sales team and an online education sales team.


\subsection{Interface}
\label{sec:interface}
\modelname system is a web application which is deployed on the Tencent cloud and can be easily integrated in the customers' CRM system.
\footnote{The demonstration video for \modelname can be viewed in \url{https://www.youtube.com/watch?v=87rxct_HUjU}.}
We show the user interface and user interaction of \modelname in Fig \ref{fig:interface}. 
The users start from an entrance page that includes chatlog file uploading and task selection. After uploading the chatlog in the \textsc{csv} format, the users could select a specific task with the ``task starts'' operation ({\large \textcircled{\small 1}}), and view the details of the finished tasks in the ``task list'' section.
The selected task in the entrance page corresponds to different modules of the system ({\large \textcircled{\small 2}}-{\large \textcircled{\small 5}}).
For the \textit{customer FAQ extraction} module ({\large \textcircled{\small 2}}), the extracted FAQs would be plainly shown in a new webpage that does not need further user interaction.
For the \textit{customer objection response} module, the response mining sub-module corresponds to the information table of the clustered customer messages ({\large \textcircled{\small 3}}), the users could refer to the clustered sales responses by clicking on a specific message cluster in a popup window ({\large \textcircled{\small 6}}). The user could also gain sales insights by searching for the existing sales responses in the searchbar ({\large \textcircled{\small 4}}).
On the \textit{sales manager dashboard} page ({\large \textcircled{\small 5}}), the sales managers could overview the execution ratio of the sales SOP from different viewpoints, i.e. trigger view, team view and staff view, by interacting with the tabbar.

\begin{figure}[t]
	\centering
    \includegraphics[width=1.0\linewidth]{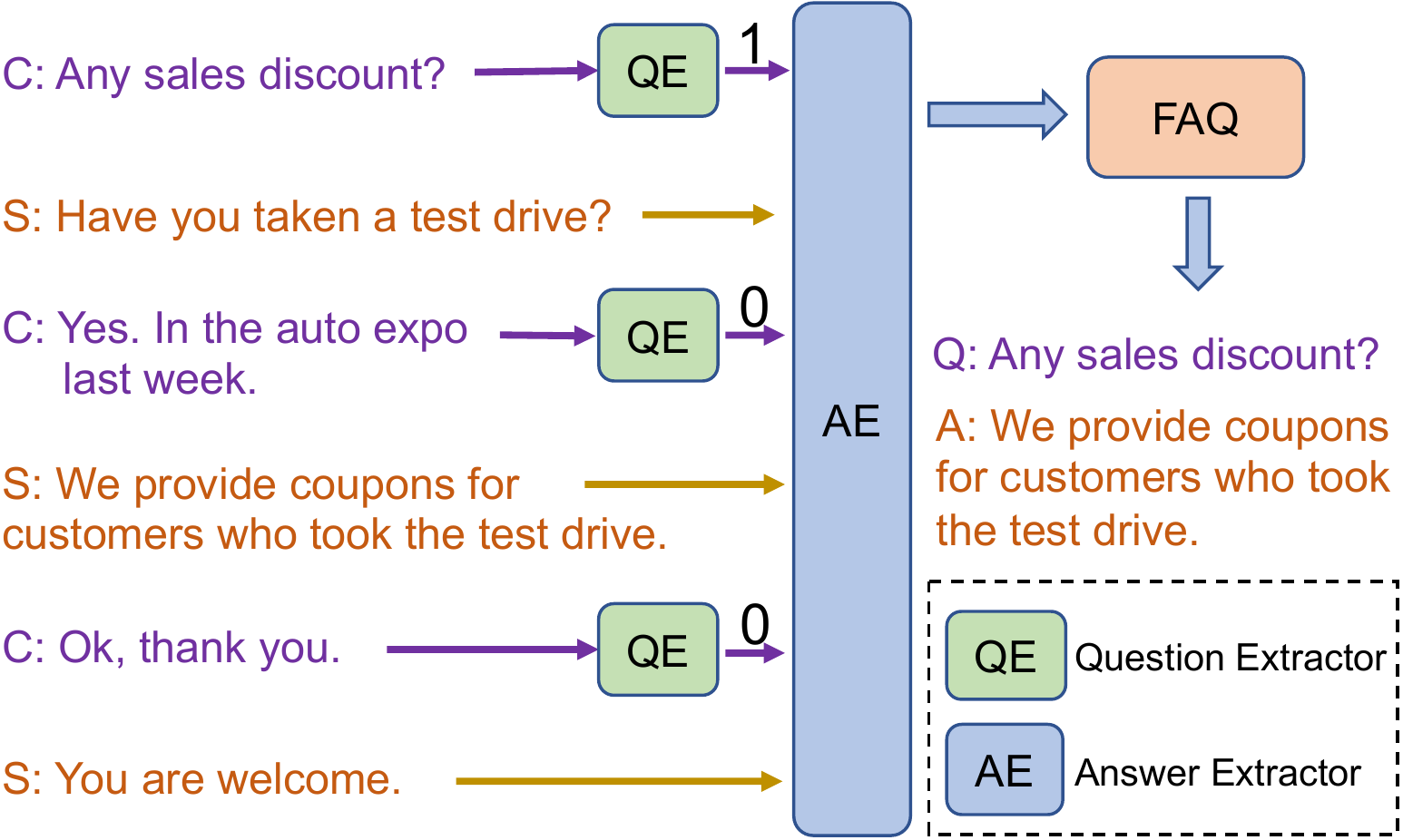}
	\caption{The workflow of the \textit{customer FAQ extraction} module.}
	\label{fig:faq_arch}
\end{figure}

\section{AI Engine}
\label{sec:ai_engine}
\paragraph{QA pair mining}
As depicted in Fig \ref{fig:faq_arch}, the QA pairs mining module is comprised of a question extractor and an answer extractor. The question extractor aims to determine the \textit{meaningful} customer queries and is formulated as a multi-label classifier for \textit{semantic integrity}, \textit{chitchat filtering}, \textit{legal customer inquiry}\footnote{A valid question should be labeled as ``yes'' for all three aspects.}, which is modeled as a one-layer MLP classifier over the \textit{CLS} vector of a syntax-enhanced BERT \cite{li-etal-2021-improving-bert,xu-etal-2021-syntax} encoder.
We model the answer extractor with a prompt-based BERT matching model after recognizing the valid customer queries. The input of the answer extractor is a dialog snippet $\mathcal{S}=\{u_1^q, u_2, \dots, u_r\}$ with $r$ utterance, in which $u_1^q$ denotes the valid customer query recognized by the question extractor while $\{u_2, \dots, u_r\}$ denotes the subsequent utterances that includes the messages from both customer and sales. We feed the dialog snippet $\mathcal{S}$ with soft prompt templates \cite{qin-eisner-2021-learning,hambardzumyan-etal-2021-warp,schick-schutze-2021-exploiting} into a BERT encoder. For simplicity, suppose there only exists 4 utterances in $\mathcal{S}$, the input format is 
$
{\rm BERT}(\{ {\rm [C]} \, u_1^q \, {\rm [S]} \, {\rm [V_1]} \, u_2^a \, {\rm [S]} \, u_3^q \, [S] \, {\rm [V_2]} \, u_4^a \, {\rm [S]}\})
$,
where ${\rm [C]}$, ${\rm [S]}$, ${\rm [V_1]}$, ${\rm [V_2]}$ represents the \textit{CLS} token, the \textit{SEP} token, and the prompt template tokens that are inserted before the candidate answers \{$u_2^a$, $u_4^a$\}, respectively.
Then we derive the answer scores
$
s_k^a = {\rm sigmoid}({\rm MLP}(h^{\rm [V_k]}, h^{\rm [C]}))(k=1,2)
$ for \{$u_2^a$, $u_4^a$\}. 
We select the answer with the highest score for $u_1^q$ only if the score is larger than a threshold, i.e. 0.75, otherwise we posit no valid answer exists in the chatlog for $u_1^q$.
Offline experiments show that the proposed QA extractor outperforms multiple QA extraction baselines \cite{devlin-etal-2019-bert,jia2020matching}.
The training data for QA extraction are annotated by the crowdsource workers on an internal platform of Tencent comparable to AMT.

\paragraph{Utterance Semantic Clustering}
After filtering trivial information in the dialog (Fig \ref{fig:obj_arch}), we represent the utterances with the dense vectors using the pretrained sentence encoders \cite{reimers-gurevych-2019-sentence} which are compatible with efficient query retrieval on GPUs \cite{johnson2019billion}. 
Then we assemble the utterances that are similar in the semantics with the K-means algorithm.
In each assembled cluster, we set the semantic centroid query as the anchor and use a dedicated BERT sentence-pair matching model \cite{devlin-etal-2019-bert} that is pretrained on an internal query-query matching corpus to filter the utterances with the low relevance to the anchor query in the cluster. 

\paragraph{Keywords and Semantic Classifier}
We use the open-source \textit{topmine} \cite{el2014scalable} toolkit to extract keywords from customer or sales utterances in the \textit{customer objection response} module.
For the trigger and spotlight detection in the \textit{sales manager dashboard} module, the intent classifier for a specific sales groups is tailored and has a domain-specific intent vocabulary. For example, for the sales teams in the online education domain, the pre-defined intents include ``affordability'', ``competitors'', ``eagerness to learn'', ``curriculum'' , ``lack of time'', etc.
The intent classifier also supports domain-specific keyword matching, e.g. ``new energy car'', ``driver asisstance'', ``intelligent vehicle'' for the car sales teams.

\section{Conclusion}
We propose the \modelname system to automatically extract and analyze winning sales scripts from the enormous customer-sales chatlog using advanced NLP technologies for both sales representatives and managers.
\modelname consists of three modules: 1) \textit{customer FAQ extraction} aims to enrich the FAQ knowledge base by mining question-answer pairs from chatlog; 2) \textit{customer objection response} helps the sales to figure out the typical customer objections and the best practises to respond; 3) \textit{sales manager dashboard} depicts the performances for different sales staffs or teams.
The system has been running on the Tencent Cloud for several sales teams from different domains. 

\bibliographystyle{ijcai}
\bibliography{ijcai}

\end{document}